# Multi-level biomedical NER through multi-granularity embeddings and enhanced labeling

Fahime Shahrokh [1], Nasser Ghadiri [1], Rasoul Samani [1], Milad Moradi [2]

**Abstract**

Biomedical Named Entity Recognition (NER) is a fundamental task of Biomedical Natural Language Processing for extracting relevant information from biomedical texts, such as clinical records, scientific publications, and electronic health records. The conventional approaches for biomedical NER mainly use traditional machine learning techniques, such as Conditional Random Fields and Support Vector Machines or deep learning-based models like Recurrent Neural Networks and Convolutional Neural Networks.

Recently, Transformer-based models, including BERT, have been used in the domain of biomedical NER and have demonstrated remarkable results. However, these models are often based on word-level embeddings, limiting their ability to capture character-level information, which is effective in biomedical NER due to the high variability and complexity of biomedical texts. To address these limitations, this paper proposes a hybrid approach that integrates the strengths of multiple models.

In this paper, we proposed an approach that leverages fine-tuned BERT to provide contextualized word embeddings, a pre-trained multi-channel CNN for character-level information capture, and following by a BiLSTM + CRF for sequence labelling and modelling dependencies between the words in the text. In addition, also we propose an enhanced labelling method as part of pre-processing to enhance the identification of the entity's beginning word and thus improve the identification of multi-word entities, a common challenge in biomedical NER.

By integrating these models and the pre-processing method, our proposed model effectively captures both contextual information and detailed character-level information. We evaluated our model on the benchmark i2b2/2010 dataset, achieving an F1-score of 90.11. These results illustrate the proficiency of our proposed model in performing biomedical Named Entity Recognition.

## 1. Introduction

In recent years, there has been a significant increase in the number of data mining studies to improve health-care and reduce costs. These methods use biomedical data, including electronic health records, clinical texts, biomedical articles, and laboratory reports. Managing and analyzing the large amounts of biomedical data collected in unstructured textual biomedical databases requires specialized knowledge and tools, which makes biomedical text mining and Natural Language Processing (NLP) research more critical. A key step in text mining, NER can improve the use of biomedical data by identifying and classifying biomedical entities from texts. The extracted information can be used for a variety of downstream biomedical NLP tasks [1], [2], [3].

The NER task was first developed in the general domain and then applied to biomedical texts. There are differences between the identification of generic entities and medical entities. In the general domain, entity sets are relatively heterogeneous, ranging from names of people and places to monetary amounts. In contrast, in the biomedical domain, entity sets are often limited to medically relevant terms, such as names of proteins, drugs, symptoms, and treatments. Also, there are associated resources in the biomedical that computer linguists can use

[1] *Department of Electrical and Computer Engineering, Isfahan University of Technology, Isfahan 84156-83111, Iran*
(e-mails: fahimeshahrokh@alumni.iut.ac.ir, nghadiri@iut.ac.ir, rasoul.samani@ec.iut.ac.ir)

[2] *Institute for Artificial Intelligence, Center for Medical Statistics, Informatics, and Intelligent Systems, Medical University of Vienna, Vienna, Austria* (e-mail: milad.moradivastegani@meduniwien.ac.at)

Corresponding Author: Nasser Ghadiri
Affiliation: *Department of Electrical and Computer Engineering, Isfahan University of Technology, Isfahan, Iran*
(email: nghadiri@iut.ac.ir)

as a reference to identify the famous entities of this domain. Medical phrases may have various writings, synonymous terms and terminology in this domain that complicate text processing. Many named entities contain more than one word, which makes identifying the boundaries of multi-word expressions more challenging. It is common for physicians to write clinical notes in different ways, often abbreviating them and using poor grammar or structure. These and other complexity and heterogeneity make it challenging to identify named entities from biomedical texts, particularly clinical reports [4], [5], [6], [7].

Various approaches to biomedical NER have been proposed. These methods range from conventional methods, such as rule-based and dictionary-based techniques, to more advanced ones, such as machine learning-based, deep learning-based, and transfer learning-based methods.

Dictionary-based or rule-based NER approaches use existing vocabularies or knowledge sources (such as UMLS [8] or SNOMED [9]) to build a collection of terms or rules. These terms are tagged in the text using string exact matches or variations that comply with the defined regulations. Even though this approach is simple, it is limited in its ability to generalize to unseen data, coverage of the rules and the complexity of the biomedical domain. It also requires significant human effort to develop a vocabulary and rules [10].

The shortcomings of traditional NER methods have been overcome by supervised machine learning, including conditional random fields (CRFs), support vector machines (SVMs), and logistic regressions [11]. SVMs are widely used in for text classification problems, including NER and are effective in handling high-dimensional feature spaces [12]. CRFs, on the other hand, are a popular machine learning approach that takes into account the context of words in a sequence and can effectively identify named entities such as drug names and features in clinical notes [13]. While machine learning has improved NER significantly, they still rely heavily on hand-crafted features and human intervention for feature engineering [14].

Deep learning-based methods, such as Recurrent Neural Networks (RNNs) [15] and Convolutional Neural Networks (CNNs) [16] have gained widespread popularity in recent years due to their ability to capture complex patterns in data. The automatic feature engineering and self-learning capability of deep learning algorithms make them less reliant on human intervention compared to traditional machine learning algorithms. While deep learning methods have demonstrated strong performance in NER tasks, including biomedical NER, they also require large amounts of labeled training data and are computationally expensive [17].

Recently, research in NLP has shown that the use of pre-trained language models on unannotated large text datasets, such as the BERT [18] as a text representation model, will significantly improve the performance of NER tasks on smaller or domain-specific datasets [19]. The advantage of transfer learning-based approaches for biomedical NER is that they can overcome the challenge of limited annotated data in the target domain. By exploiting PubMed and MIMIC-III [20] datasets, various language representation models have been developed for representing medical and clinical text. The BLUE [21], BioBERT [22], and ClinicalBERT [23] are among the models used to facilitate the extraction and processing of clinical concepts.

Furthermore, hybrid methods have been proposed for biomedical NER in addition to these individual approaches. These methods use machine learning or deep learning as the base model and enhance performance by leveraging additional information, such as dictionaries or combining multiple models [24].

According to the mentioned shortcomings of existing approaches, in this paper, a multi-layered neural network architecture has been proposed that uses representations, language models, and embeddings at the character and word level to identify Named Entities. The model is designed to consider various aspects of word meaning, context, structure, and writing format. It utilizes pre-trained biomedical BERT models to create word-level embeddings, which help capture the meaning and context of words. Additionally, the model uses a CNN network to identify character patterns of words to improve the detection of out-of-vocabulary words in the word embedding model. Furthermore, the paper presents an enhanced labeling method as part of the preprocessing stage, which improves the identification of multi-word entities. The effectiveness of this labeling method on the model's performance is also evaluated. Word embeddings are highly effective, but in some cases, they are not enough and are not appropriate in all instances, such as out-of-vocabulary words, spelling mistakes, and different spelling forms of the same entity. To address these limitations, a character-level structure can be considered to enhance model performance [25]. By utilizing a CNN model and processing words at the character level, patterns within words can be discovered [26]. Additionally, Writing Format Embedding can aid in recognizing specific biomedical terms that are often written in a particular format, including acronyms and abbreviations [27]. Our proposed model utilizes a Bi-directional Long Short-Term Memory (BiLSTM)+CRF [28] architecture for sequence labeling, as this architecture has demonstrated effectiveness in NER tasks by modeling term-to-term dependencies [27] [29].

Our approach combines the advantages of these elements to improve the performance of biomedical NER. In addition, we propose an enhanced labeling method that improves labeling performance compared to other works. Our experimental results on a benchmark biomedical dataset demonstrate that our proposed model has higher performance than its competitors in F1-score.

The contributions we make in this paper are as follows:

- Designing a hybrid model for processing each word in various aspects of semantic relationships, structural features, and writing format
- Using word-level embedding based on pre-trained biomedical BERT models in order to capture the meaning and context of the words
- Using character-level embedding with CNN network to discover character patterns of words in order to improve the detection of out-of-vocabulary words in the word embedding model.
- Performing extensive experiments to evaluate our model with multiple biomedical pre-trained embedding models and investigating the impact of the word embedding model on model performance.
- An enhanced labeling method is presented as part of the preprocessing stage in order to improve identifying multi-word entities and evaluate its impact on the model's performance.

The rest of this paper is organized as follows: Section 2 reviews related work on NER and existing approaches in this field. Section 3 then introduces our proposed model, detailing its different components. In Section 4, we demonstrate our model's effectiveness for NER through extensive experiments. Finally, Section 5 concludes with a summary and discussion of future work.

2. **Related Work**

In this section, we will briefly review these approaches and highlight their strengths and weaknesses.

- **Dictionary and Rule Based**: Zhang et al. [30], proposed an unsupervised solution using the inverse document frequency (IDF) method. To determine the candidate's words, they use the IDF technique to score the words in the data set. Then, using the UMLS knowledge set, the words' similarities are evaluated, and the words with the highest similarity are considered named entities. Learning without supervised supervision and labeling is an advantage of this model. In the study [3] an unsupervised model is proposed using UMLS and MIMIC-III dataset. To begin, FastText [31] is applied to the MIMIC-III dataset to generate a word embedding model. This word embedding model produces the word features vector of the input words, which is then compared to the UMLS concepts feature vector using the Cosine similarity criterion. Phrases are classified as named entities if the two feature vectors are similar.
- **Machine Learning Based**: The benefit of SVMs is that they can deal with high dimensional feature spaces, and dense feature vectors, making them ideally suited to text classification tasks [32]. Roberts et al. [33] have proposed a model to use support vector machines with manual features to extract anatomical location data related to the radiology reports. Sarker et al. [34] proposed an automatic text classification method for detecting drug side effects from texts in social networks such as Twitter using three supervised classification approaches: Naïve Bayes (NB), SVM and Maximum Entropy (ME). They showed that among the three classifiers, SVMs performed significantly better than the other two. Logistic Regression is another machine learning algorithm, which has often been used to detect entities and their relationships in text data. Rochefort et al. [35] have used multivariate logistic regression to detect events and relationships between them from electronic health record notes. The CRF model is another widely used algorithm that has been used in many articles to recognize named entities [36].
- **Deep Learning Based:** Recent research on text representation and text classification has used deep learning algorithms such as CNNs and RNNs. Hofer et al. [37] have proposed a hybrid model based on character and word embedding, which uses the CNN model in the character-based embedding and the GloVe [38] in the word embedding. The written format of words was also used as an auxiliary embedding by Chiu [27]. Li's study [39] embeds words by using the GloVe model, which is then processed by BiLSTM networks. Furthermore, Ji et al. [40], proposed a model based on LSTM and the CRF and the performance of different LSTM structures in recognizing named entities has been evaluated. The results of this research show the better performance of Multi-Layer Fully-Connected LSTM compared to other types of LSTM models. Tang et al. [41] proposes a model based on CNNs and word embeddings, which classifies text and sentences. Text sentence vectors with candidate windows, have been represented with CNN models.

- **Transfer learning-based**: Lee et al. [22], introduced the BioBERT model as a pre-trained language representation model for the medical domain and trained on data from the biomedical domain (abstracts of PubMed articles and full text of PMC articles). BioBERT evaluations show the improvement of model performance in question answering and NER tasks compared to previous works. Huang et al. [23] presented the ClinicalBERT model which trained on the MIMIC-III dataset, which consists of clinical reports of patients for use in clinical text mining. The BLUE model [21] is another model based on the BERT model which is pre-trained on the datasets of PubMed article abstracts and MIMIC-III clinical reports. In Lewis et al. [42] work, several BERT-based models trained on biomedical domain data have been evaluated in several biomedical and clinical text mining tasks. The investigations show that the BioBERT model in biomedical text mining and the RoBERTa-Large [43] model in clinical text mining have performed better than other models. In [44], Si compares traditional word representation methods, such as Glove and FastText, with advanced contextual representation methods like ELMO and BERT. Their results highlight the benefits of embeddings achieved through unsupervised pretraining on clinical text corpora like MIMIC-III. These embeddings achieve higher performance than off-the-shelf embedding models and result in a new state-of-the-art performance for the NER task. Yang's study [45] aims to evaluate the performance of transformer-based models for clinical concept extraction with pre-trained clinical models to facilitate clinical concept extraction and other clinical NLP tasks. Nath et al. [24] employed a combination of pre-trained language models and deep neural networks to recognize named entities. They proposed three network architectures based on BiLSTM-CRF networks for identification, and ClinicalBERT for generating word vectors. One significant advantage of their model is its proficiency in detecting polarity and negation. Bhattacharya et al. [46] employed transfer learning with a BiLSTM-CRF model. It was first pre-trained on a large named entity recognition corpus and used an asymmetric tri-training methodology for biomedical named entity recognition. The methodology incorporated three models: BiRNN-CRF, BiLSTM-CRF, and BiGRU-CRF. These models were then combined in an ensemble to predict the data, resulting in enhanced performance. Banerjee et al. [47] have reformulated the task of NER as a multi-answer knowledge-guided question-answering (KGQA) problem. Their approach involves prepending five different contexts, including entity types, questions, definitions, and examples, to the input text. They then train and test BERT-based neural models on these sequences from a combined dataset. Pavlova et al. [48] also propose a novel method for pre-training language models. They introduce "contextualized weight distillation," which leverages BERT's ability to create domain-specific word embeddings. Four models were pre-trained using different methods, including training from scratch, continuation from BERT-base, averaging token weights, and utilizing the proposed contextualized weights. The advantage of their work lies in the increased training speed of pre-trained models and the improved accuracy of the NER task.

After reviewing the current approaches in named entity recognition, it can be concluded that dictionary-based models have limited accuracy and vocabulary. Machine learning methods require significant manual feature engineering, resulting in increased model development time. Deep learning approaches, while highly effective, demand abundant labeled resources to train models. These findings highlight the need for further advancements in developing accurate and efficient named entity recognition models. In this paper, we have proposed a BERT-based model specifically trained on a biomedical dataset and character-level input to address the issue of out-of-vocabulary words. The proposed model employs multiple inputs to analyze the semantic aspects of medical terms and accurately extract concepts, which helps overcome the challenges in this field. To design a high-performance model, we have incorporated the capabilities of CNN and Bi-LSTM networks, such as the discovery of word character patterns in the character-level representation of words and process sequential data bidirectionally, which means that they can take into account both the preceding and succeeding words when predicting the current word's label.

## 3. Methods

Our proposed model for named entity recognition integrates multiple techniques for effective text analysis. The model is composed of several components, each with a specific role in detecting and labelling entities within the input text. Firstly, a fine-tuned BERT model is utilized to generate word embeddings, which provide a comprehensive representation of the contextual information associated with each word. Secondly, we use a multi-channel CNN to produce character embeddings, which enable the model to capture finer-grained information about the input text, such as word morphology and spelling variations. To further enhance the performance of the model, we incorporate writing format embeddings, which allow the model to identify specific patterns and structures within the text that are indicative of certain entity types. Additionally, we employ a BiLSTM+CRF layer for sequence labeling, which enables the model to capture dependencies between words in the input text and produce more accurate entity labels. Finally, we utilize an enhanced labelling method to improve the quality of the labelling process, ensuring that the model is able to correctly identify and label entities within the input text with a high degree of accuracy. By combining these various techniques, our hybrid model is able to effectively capture both word-level and character-level features, leading to better performance in named entity recognition tasks.

### 3.1. Character Embedding

In our modeling process, we consider the character-level attributes of words and discover their corresponding character patterns. By producing closely related property vectors for words that have the same source and refer to the same concept, the final model will better overcome the challenge of words out of vocabulary. To extract comprehensive character-level features [16], We employ a multi-channel one-dimensional CNN architecture with three parallel channels. Each channel has a distinct filter size, measuring 3, 5, and 7. An overview of the architecture is shown in Figure 1.

The input to each channel comprises the character embedding of one word. To prepare data for this CNN model, we use a reference array of 97 elements to map the characters of words to their corresponding numerical vectors. This array contains lowercase and uppercase English characters, numbers, and distinct characters. The input data needs to be a numerical vector of fixed length. The objective is to obtain a vector representation for each word while avoiding word splitting. To achieve this, the length of the longest word in the dataset is used to determine the size of the words' character mapping vector. Zero-padding is applied to guarantee that all word vectors have the same length.

The 1-D convolution layer extracts features when performed on the input word using the corresponding filter. 3×1 filters capture local features or short-term dependencies between characters, for instance, certain letters or letter combinations, which help identify infrequent words, such as test factors or drug names. The 5x1 filters capture medium-term dependencies that enable the system to recognize common patterns in terms of text, such as prefixes and suffixes. The 7×1 filters help capture global features or long-term dependencies that contribute to identifying the overall structure of words, including their length and composition. After the convolution layer, a max-pooling layer is applied to the feature maps produced by each channel to extract the most significant feature for each position. The resulting feature maps from all channels are concatenated and processed through a fully connected layer to generate a fixed-length feature vector for each token in the input sequence. To avoid over-fitting, we also applied a dropout layer at the end of the output of the Flatten layer [37]. This feature vector serves as the character-level representation for the word in our final NER model.

### 3.2. Word Embedding

The model's architecture is designed so that, in addition to traditional word representation models such as GloVe, new word embedding models such as BERT and ELMo [49] can also be used in the input part word embedding. When these models are trained with large amounts of biomedical data and texts, they are more likely to learn a wide range of specific expressions in the field. In clinical texts, BERT-based models can be used to overcome a lack of clinical textual resources and the presence of field-specific terms and by using these textual representation models, the training process becomes less time-consuming. Word embedding models generate a numerical vector for each word unit in a text by considering the unit's position in the text containing it and the words surrounding it. As a result, the input of these models is a text containing words.

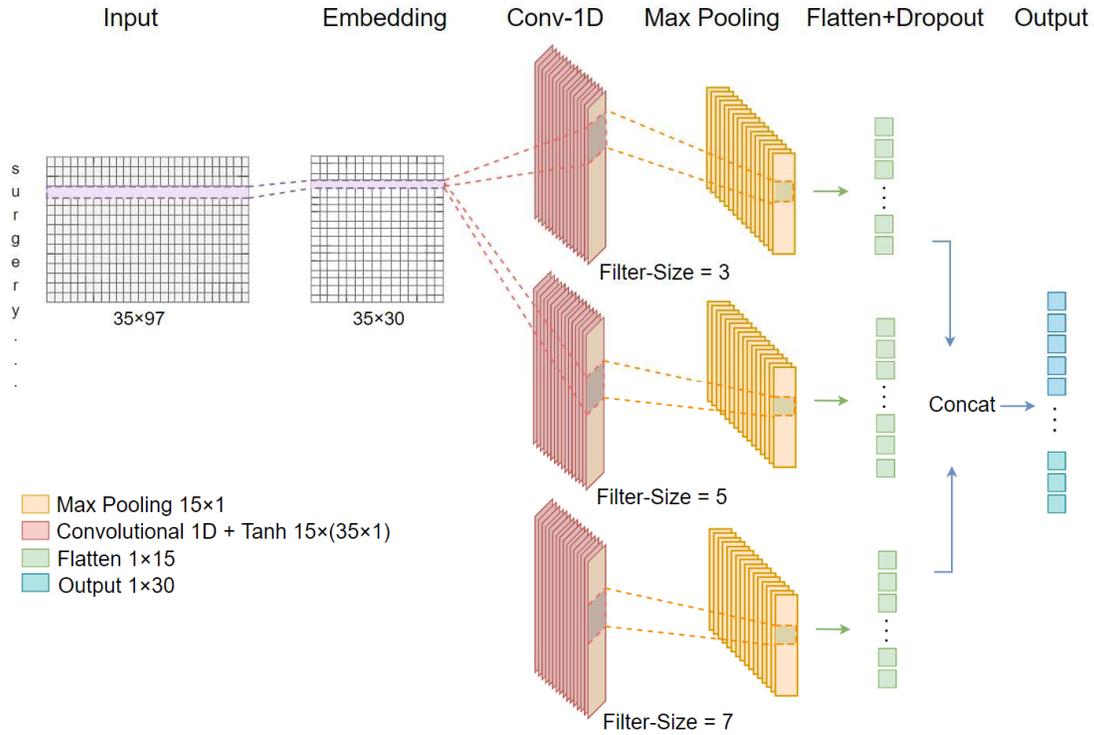

**Figure 1.** Multi-Head CNN model architecture for character embeddings. The model comprises an embedding layer of word vectors. A Convolutional 1D layer with three filters size is then applied, followed by max pooling. The output is obtained by flattening the layer and applying dropout regularization.

In order to use BERT-based models in word embedding input, the first step is to perform a fine-tuning process on these models so that the model is trained and updates its weights based on the Token Classification process. During the fine-tuning process, different models built on the BERT base model and trained on data from the biomedical domain have been used. The weights obtained from fine-tuning these models are used in the word embedding input of our model.

After fine-tuning the various BERT-based models, it is time to receive text embeddings from these models. There are several methods for obtaining embeddings from BERT models. Considering that the BERT architecture is a 12-layer architecture, each layer can produce an embedding related to the input. Using the last hidden layers of this model is recommended to get proper embedding because the last layers have more details and optimal features than the bottom layers. Because BERT models are fine-tuned in the word embedding section, the last hidden layer of each model is used as the embedding generation layer [50]. There are several methods to generate an embedding of a word in BERT models. After evaluating all the methods to increase the model's efficiency in the proposed model, the average of sub-tokens embedding has been used to generate word embedding.

### 3.3. Writing Format Embedding

One of the complicated aspects of biomedical texts, especially clinical reports, is the extensive use of technical terms and non-standard and informal abbreviations. This has made it difficult for non-specialists in this field to understand clinical reports and systems for extracting knowledge from the text [27]. As mentioned in the related work section, using word embedding models previously trained on a large set of texts in biomedicine increases the probability of recognizing words specific to this field, including these modifications and abbreviations. Another approach to overcome the challenge and improve the model's performance is to embed the written format of the word and use it as an auxiliary input of the model.

The input of this step is each word in the dataset. Its output is a numerical vector with a length of eight, specifying the correct written format of the word. Finally, the vector resulting from embedding the word in the writing format is used as one of the inputs for the aggregation layer of the model. An example of a named entity and the vector corresponding to its writing format is shown in **Error! Reference source not found.**.

```
                            a          [0 .0 .0 .0 .0 .0 .1 .0]
a c5-6 disk herniation  →   c5-6       [0 .0 .0 .0 .0 .1 .0 .0]
                            disk       [0 .0 .0 .0 .0 .0 .1 .0]
                            herniation [0 .0 .0 .0 .0 .0 .1 .0]
```

**Figure 2.** A sample for writing format embedding

Writing format input is implemented to help recognize words with specific spellings in text. From the examples of words with exceptional writing, we can refer to words that start with capital letters or words that have numbers as part of their characters. This type of word writing is used in the abbreviation of texts and particular words in medicine. an array including eight written states of words is considered in this part of the model.

### 3.4. Input Integration and Label Prediction

Given the ability of RNN models to preserve long-term dependencies in data sequences, the proposed model uses the BiLSTM model as an RNN model with the ability to capture features in both directions, forward and backward. In the last step of our proposed method, shown in Figure 3, we use a two BiLSTM layers architecture, followed by a CRF layer for sequence labelling. The input for the first BiLSTM layer is the concatenation of word embeddings generated by the fine-tuned BERT model, the character-level embedding created by the three-channel CNN, and writing format representations of words. First BiLSTM layer processes this input and makes hidden states for each word, which capture the semantic dependencies between the words within the input text. The second BiLSTM layer captures longer word dependencies by processing the output of the first BiLSTM layer. Using the second BiLSTM layer enhances the capability of the ultimate model for identifying named entities containing multiple words. Following the second BiLSTM layer, a Softmax layer sets scores for each class, representing the likelihood that each word of the text belongs to a specific class. After calculating the scores, the CRF layer models the dependencies between labels and generates a final label for each word. The CRF layer allows the model to make more accurate predictions by considering the label arrangement for the named entities. For example, label beginning with "I" do not appear at the start of the name of a named entity. Labels starting with "I" appear only before words labeled "B" or "I".

### 3.1. Enhanced Labeling

We used the IOB (Inside-Outside-Beginning) standard for tagging. In this method, the first word of the named entity receives the initial label of the phrase with the prefix "B", followed by subsequent words in the phrase receiving the label with the prefix "I", and other words receiving the label "O" [51]. We used the CRF model, a popular labelling model used in NLP tasks. The CRF model recognizes named entities by applying rules learned from word labels during the model training process. In this model, the label for the present word is based on the label for the previous word. Identifying the first word in a multi-word concept with a label "B" will increase the likelihood that subsequent words with labels "I" will be identified [52] [53]. There are some multi-word named entities in the i2b2 2010 dataset with stopwords and frequent words as the first word labelled "B". Pre-trained word embedding models like BERT have often viewed these words as unlabeled terms during the training process, so our model also considers these words with the label "O".

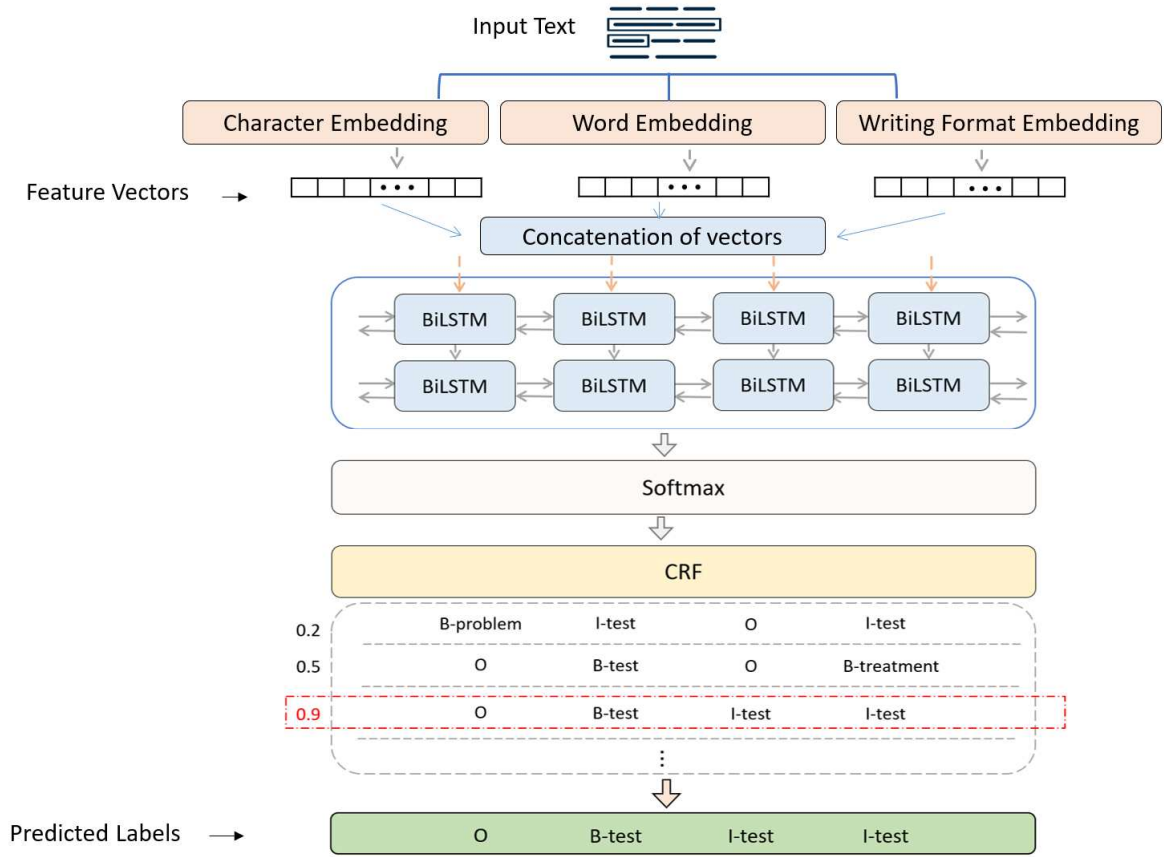

**Figure 3.** The sequence of predicted tags with the highest probability in the CRF layer is selected as input word tags

As a result, the model's efficiency decreases when the probability of identifying such multi-word concepts decreases. To avoid this issue, we identify stopwords and frequent words with the label "B". If they are not biomedical abbreviations and their next word label is "I", the first word's label is changed to "O", and the next word's label is changed to "B". Some examples of this change in labelling demonstrate in Table 1.

**Table 1.** An example of enhanced labelling that changes the labels of the stopword "a" and the frequent word "patient" at the beginning of a named entities

|  | Sample | Description |
| --- | --- | --- |
| Phrase | a bacterial superinfection |  |
| Phrase Initial labelling | [problem] |  |
| Word's IOB labelling | [B-problem] [I-problem] [I-problem] | At the beginning of the phrase, the label of the stopword "a" changes from "B" to "O", and the label of " bacterial " changes from "I" to "B". |
| Word's Enhanced labeling | [O] [B-problem] [I-problem] |  |
| Phrase | Patient 's neurologic exam |  |
| Phrase Initial labelling | [test] |  |
| Word's IOB labelling | [B-test] [I-test] [I-test] [I-test] | Labels of the commonly used word "patient" and Possessive "s" changes from "B" to "O" at the beginning of named entities. And the label of " neurologic " changes from "I" to "B". |
| Word's Enhanced labeling | [O] [O] [B-test] [I-test] |  |

## 4. Experimental Evaluation

In this section, we present a comprehensive experimental evaluation of our proposed biomedical named entity recognition (NER) model. We aim to demonstrate the effectiveness and robustness of our approach in accurately identifying and classifying biomedical entities within input data. To achieve this, we have structured our evaluation into these subsections: Dataset, Enhanced Labeling, Fine-tuning Pre-trained Model, and Result Comparison.

### 4.1. Dataset

We used the i2b2 2010 [54] dataset to evaluate the performance of our model in identifying and extracting concepts from biomedical texts. This dataset contains 170 patient reports for the Training set and 256 reports for the Test set. Concept extraction for this dataset involves identifying named entities from patient reports. Named entities in the dataset reports are annotated with three labels: Problem, Treatment, and Test. Table 2 shows an analysis of the number of named entities in the test and training texts.

Table 2. The number of named entities in the test and training sets

| Entity | Test set | Training set |
| --- | --- | --- |
| Problem | 12592 | 7073 |
| Test | 9225 | 4608 |
| Treatment | 9344 | 4844 |

### 4.2. Enhanced Labeling

As mentioned in Section 3.1, in order to improve the accuracy of the model, phrases whose initial word is not a biomedical abbreviation and their next word label is "I" will have the first word's label changed to "O", and the next word's label will be changed to "B". According to our study of some biomedical references, some words in the set of stopwords appear as abbreviations in biomedical texts, especially in laboratory results. Therefore, we removed these stopwords from the previous preprocessing task and retained their original labels. Table 3 shows the frequency of labels in the training and test sets according to our enhanced labelling method. Among the named entities, "problem" has the highest frequency, and "test" has the lowest.

Table 3. Percentage Distribution of Classes in Train and Test Sets: Before and After Enhanced Labeling

|  | Before enhanced labelling | | After enhanced labelling | |
| --- | --- | --- | --- | --- |
| Label | Test set | Training set | Test set | Training set |
| B-problem | 4.95% | 4.97% | 4.96% | 4.97% |
| I-problem | 6.96% | 7.20% | 6.15% | 5.88% |
| B-treatment | 3.68% | 3.40% | 3.68% | 3.41% |
| I-treatment | 3.13% | 2.88% | 2.52% | 2.39% |
| B-test | 3.63% | 3.23% | 3.64% | 3.24% |
| I-test | 3.15% | 2.73% | 2.32% | 2.60% |
| O | 74.48% | 75.56% | 76.74% | 77.51% |

Continuing with the data analysis procedures, we show the distribution of the number of words in each named entity in dataset. Figure 4 provides valuable insights into the characteristics of the named entities in i2b2 dataset. It reveals that the majority of named entities are one, two or three words in length, accounting for over 80% of the total entities identified. Moreover, the high number of multi-word entities emphasizes the importance of identifying them in order to increase model efficiency. As we analyzed the length distribution of the named entities in this dataset, we tried to improve the final model's ability to recognize multi-word entities.

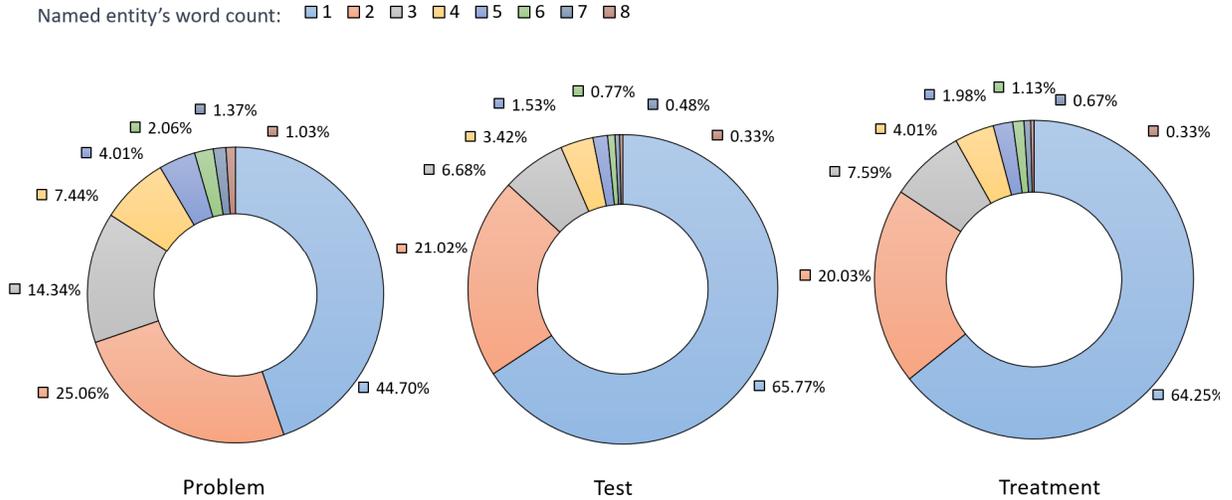

**Figure 4.** Length distribution of named entities in the dataset

### 4.3. Fine tuning of pre-trained models

In the first experiment, the efficiency of the preprocessing method is evaluated. The impact of preprocessing on the performance of the BlueBERT-base model for the NER task is demonstrated in Figure 5 through fine-tuning. As the results show, first, preprocess is removing stopwords from the text. Applying this preprocessing has significantly reduced the F1 measure for the model. Removing stopwords as a part of the text components, which in most cases establish the connection between other text parts, can cause the text structure to be broken, and the model's performance will decrease.

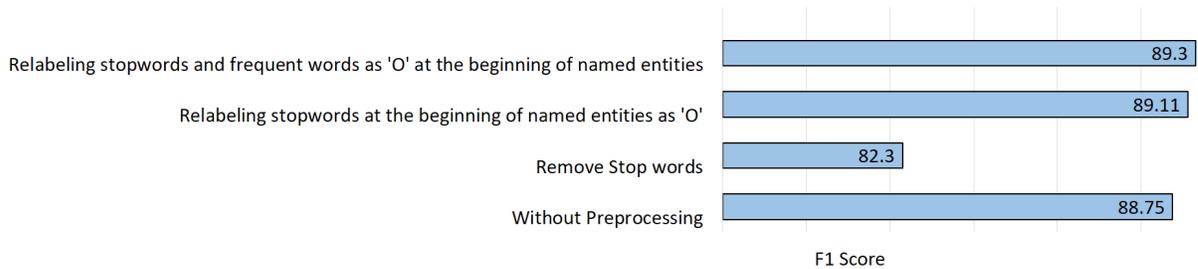

**Figure 5.** Impact of different preprocessing techniques on fine-tuning BERT model performance.

In the following, we will examine and evaluate various models based on the BERT architecture using our dataset. These models have been specifically chosen for their ability to process medical and scientific texts. Unlike general BERT models, these models have pretraining on biomedical texts and scientific articles, making them well-suited for our domain-specific tasks. The BERT-based models employed in our study include ClinicalBERT, BioBERT, SciBERT [55], BlueBERT, AlBERT [56], and RoBERTa. We perform a fine-tuning process on these pre-trained models to adapt them to our specific NLP tasks. In the following, we provide a brief overview of the training data utilized for each model.

- ClinicalBERT : This model is based on the BERT, pre-trained on the MIMIC-III clinical report dataset; therefore, this model can efficiently process clinical texts.
- BioBERT :This model is pre-trained using a collection of PubMed medical article summaries and the full text of PMC articles and is highly efficient in processing scientific medical texts.
- BlueBERT : This model is pre-trained using the MIMIC-III clinical reports dataset and PubMed article abstracts, which, compared to the ClinicalBERT model, can efficiently process biomedical texts.

- SciBERT : This model is specifically designed for the biomedical and scientific domain and has been trained on a large-scale corpus of scientific literature, including biomedical research articles, clinical notes, and other scientific texts.
- AlBERT and RoBERTa : These two models are an improved version of the original BERT model, which only differs from the original BERT model in the main structure of the model. By changing the size of the parameters and the pre-training process, these two models have been able to gain more speed in the training process and process texts with higher efficiency.

In the following, each network hyperparameter's effect on the model's performance is evaluated. Experiments have been performed repeatedly to measure the adjustment of different combinations of hyperparameters on the model's performance. In the first step of the tests of this section, the fine-tuning model implemented with initial values for the network hyperparameters was evaluated on the BERT-based models presented in the fine-tuning section. After that, in several stages, different values are set for one or more hyperparameters of the network, and the model's performance is measured.

We evaluated the model to determine the optimal value of the learning rate, utilizing different values ranging from 2e-5 to 5e-5. Among the trained models in the initial evaluation, the BlueBERT model demonstrated the highest performance. Specifically, the model achieved the highest F1 score when using a learning rate of 3e-5.

In finding optimal value for hyperparameters, the evaluation of maximum sequence length and batch size values is investigated. Figure 6 shows the effect of setting different values for these two parameters by keeping other network parameters constant, such as a learning rate equal to 3e-5 on the test dataset. For the maximum sequence length, values in the 100 to 200 were considered, and the model was evaluated.

Choosing small values such as 100 for the max sequence length due to the deletion of a part of the input text causes the loss of part of the information in it and thus reduces the model's performance. Since the maximum number of words in a sentence in the dataset was nearly 200, choosing 200 was optimal to prevent broken sentences. Of course, This parameter is considered variable due to the change of the Batch size and the limitation of processing resources.

Also, the batch size was set in several stages with values of 16, 32, and 64. Choosing a low value for this parameter has a negative effect on the training step and the model's performance due to a smaller number of samples in a batch. At the same time, valuing this parameter with large values was not possible due to the limitation of available processing resources. Setting a combination of values for these two parameters is more appropriate to obtain the highest performance measure.

The number of training epochs was also set and tested with values from 3 to 10, and increasing the number of training epochs up to a threshold positively affected the model's performance. After that, by increasing the number of training epochs, there was no significant effect on improving the model's performance. The best performance is obtained in the seventh iteration.

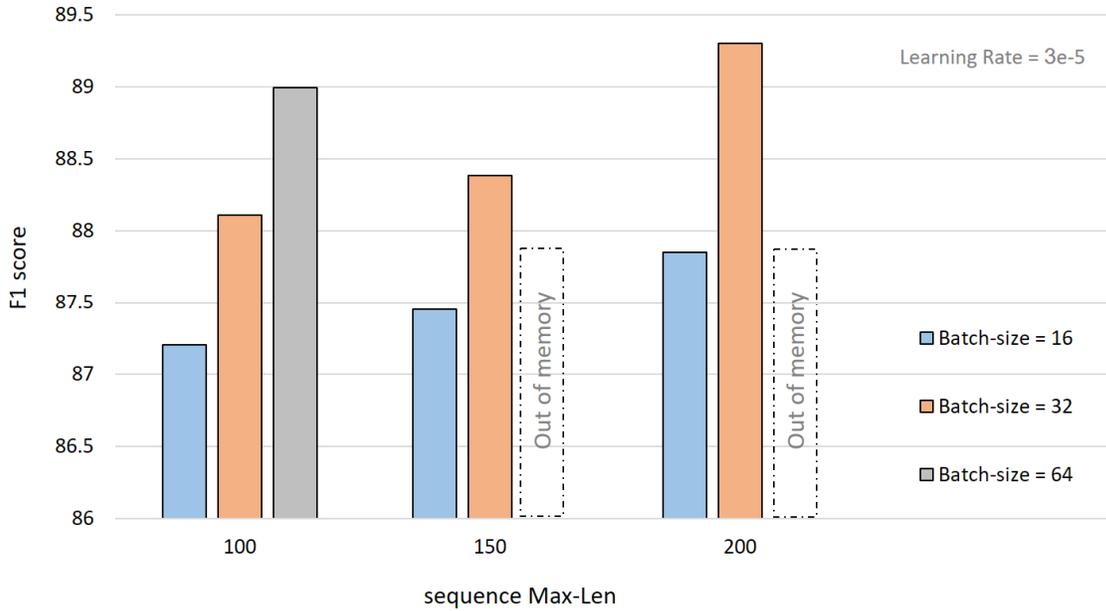

**Figure 6.** Evaluating the impact of both max sequence length and batch size parameters on model performance

In Figure 7, the process of changing the loss value during the fine-tuning process on the BlueBERT and BioBERT models is shown. As observed in this figure, for the BlueBERT model, the amount of loss in the training process on the evaluation dataset decreases until the fourth iteration, and there is not much change in this amount from the fourth iteration onwards. Also, as expected, the amount of loss follows a relatively decreasing trend in the training dataset. Starting from the fifth iteration onwards, there is no significant reduction in loss, so the training process is stopped.

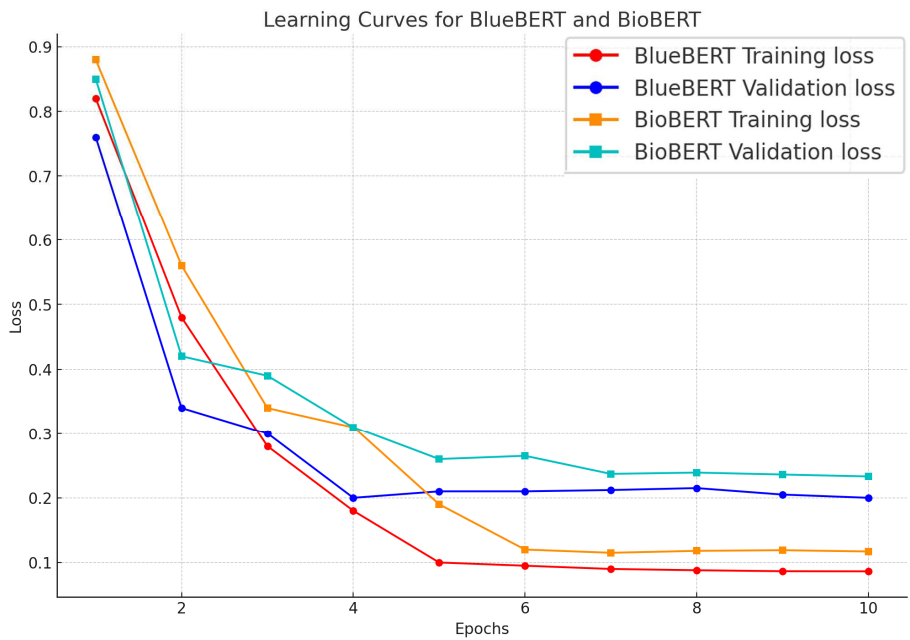

**Figure 7.** The training and validation loss curves during the fine-tuning of the BlueBERT and BioBERT models.

The optimal values were finally selected for the finetuning step are shown in Table 4.

**Table 4.** Hyperparameter setting in BlueBERT model for NER task.

| Hyperparameter | Optimal Value (values used) |
|---|---|
| Max sequence length | 200 (100, 150, 200) |
| Batch size | 32 (16, 32, 64) |
| Learning rate | 3e-5 (2e-5, 3e-5, 4e-5, 5e-5) |
| Epochs | 7 (1 To 10) |

Following the model evaluations related to fine-tuning of the word embedding model, this process has been performed on the models based on the BERT, which was introduced in the fine-tuning section. These models have been trained on general texts and medical and clinical texts. As shown in Table 5, the BlueBERT had the best performance among all the models. ALBERT, RoBERTa, and SciBERT, achieve less efficiency because they were trained on public domain texts, including Wikipedia, as expected. The ClinicalBERT is less efficient than some other models because it was trained only on the clinical report and discharge summary and did not use biomedical texts. Finally, the BlueBERT achieved the highest performance with an F1 measure of 89.30 because this model has been trained on a large set of biomedical text and clinical reports.

**Table 5.** Comparative analysis of BERT models for NER task: BlueBERT outperforms other models in performance evaluation

| Model | Class | Precision | Recall | Support | F1-Score | Average Loss | Validation Accuracy | Validation F1-Score | Validation Loss |
|---|---|---|---|---|---|---|---|---|---|
| ALBERT | problem | 0.8860 | 0.8773 | 15472 | 0.8816 | - | - | - | - |
|  | treatment | 0.8820 | 0.8897 | 14483 | 0.8858 | - | - | - | - |
|  | test | 0.8565 | 0.8553 | 10031 | 0.8559 | - | - | - | - |
|  | avg/total | 0.8771 | 0.8763 | 39986 | 0.8767 | 0.0281 | 0.9452 | 0.8774 | 0.2639 |
| RoBERTa | problem | 0.8909 | 0.8800 | 11363 | 0.8854 | - | - | - | - |
|  | treatment | 0.8854 | 0.9034 | 10531 | 0.8943 | - | - | - | - |
|  | test | 0.8492 | 0.8633 | 6817 | 0.8562 | - | - | - | - |
|  | avg/total | 0.8790 | 0.8846 | 28711 | 0.8817 | 0.0434 | 0.9590 | 0.8840 | 0.1718 |
| ClicnicalBERT | problem | 0.8857 | 0.8878 | 11198 | 0.8868 | - | - | - | - |
|  | treatment | 0.8989 | 0.8913 | 10838 | 0.8951 | - | - | - | - |
|  | test | 0.8877 | 0.8396 | 7327 | 0.8630 | - | - | - | - |
|  | avg/ total | 0.8911 | 0.8771 | 29363 | 0.8839 | 0.0267 | 0.9596 | 0.8853 | 0.1851 |
| SciBERT | problem | 0.9020 | 0.8746 | 15329 | 0.8881 | - | - | - | - |
|  | treatment | 0.9026 | 0.8919 | 13957 | 0.8972 | - | - | - | - |
|  | test | 0.8826 | 0.8351 | 9571 | 0.8582 | - | - | - | - |
|  | avg /total | 0.8974 | 0.8711 | 38857 | 0.8840 | 0.0783 | 0.9554 | 0.8887 | 0.1812 |
| BioBERT | problem | 0.8959 | 0.8867 | 11342 | 0.8913 | - | - | - | - |
|  | treatment | 0.8964 | 0.9057 | 10636 | 0.9010 | - | - | - | - |
|  | test | 0.8665 | 0.8632 | 6957 | 0.8648 | - | - | - | - |
|  | avg/ total | 0.8890 | 0.8880 | 28935 | 0.8885 | 0.0074 | 0.9602 | 0.8895 | 0.2475 |
| BlueBERT | problem | 0.9054 | 0.9010 | 14219 | 0.9032 | - | - | - | - |
|  | treatment | 0.9014 | 0.8960 | 13532 | 0.8987 | - | - | - | - |
|  | test | 0.8727 | 0.8654 | 8930 | 0.8690 | - | - | - | - |
|  | avg/ total | 0.8960 | 0.8905 | 36681 | **0.8932** | 0.0181 | 0.9524 | 0.8939 | 0.2564 |

### 4.4. Baseline Models Result

After doing experiments related to fine-tuning the BlueBERT to prepare the word-based input of the final model and obtain optimal parameters, this section will evaluate the final model's performance, and the results of the experiments will be investigated. At first, evaluation began by applying the 50-dimensional GloVe word embedding model in the word input and using a BiLSTM layer. With this structure, the model achieved F1 = 66.72.

Adding character-level embedding input increased the F1 measure by about 0.96%, and placing the writing format embedding input into the model architecture with an increase of 0.20%, bringing the model's F1 measure to 73.35%. Adding the CRF layer to the end of the model layers increased the F1 measure to 74.90%. The addition of the second BiLSTM layer increased the F1 measure by about 2%. During the experiments of this step, the proposed model was evaluated with several versions of the GloVe model with different dimensions. By increasing the epochs, the best performance achieved with this structure was F1 = 80.59%, obtained by choosing the GloVe 300.

In the following model evaluations, the testing of various word embedding models on word embedding input commenced. The performance of the ELMo model and its MIMIC-III version obtained the best performance of F1 = 88.29%. Finally, the model was evaluated using the last model obtained from fine-tuning the BERT-based models as input for word embedding. The BlueBERT achieves the best result in this input.

Furthermore, considering the selection of the BlueBERT as the word embedding input, the step-by-step experiments were repeated to add other architectural parts of the model, this time with the BlueBERT as the word embedding input. The results of these tests are given in Table 6. It should be noted that during several experiments conducted on the model, the Attention layer was used in the model's architecture. However, the use of this mechanism did not positively affect the model's performance, and one reason for this result could be the shortness of the sentences in the dataset.

By conducting step-by-step experiments and evaluating the impact of each input of the model architecture on its performance, the final model architecture consists of 1) 45-dimensional character embedding input based on CNN, 2) 8-dimensional Writing format embedding input, 3) 768-dimensional word embedding based on BlueBERT, as well as using two BiLSTM layers and the final CRF layer.

**Table 6.** An analysis of the F1-score results of the proposed model based on the addition of different components and word embedding models

| Model | F1-Score |
|---|---|
| Word(BERT) + 1-BiLSTM-CRF | 86.94 |
| Word(BERT) + Char(CNN) + 1-BiLSTM-CRF | 87.69 |
| Word(BERT) + Char(CNN) + Writing format + 1-BiLSTM-CRF | 87.83 |
| Word(GloVe) + Char(CNN) + Writing format + 2-BiLSTM-CRF | 80.59 |
| Word(ELMo) + Char(CNN) + Writing format + 2-BiLSTM-CRF | 88.29 |
| Word(BERT) + Char(CNN) + Writing format + 2-BiLSTM-CRF | 90.04 |
| Word(BERT) + Char(CNN) + Writing format + 2-BiLSTM-CRF (Ensemble) | 90.11 |

In the following, the effect of different tested values for the parameters of the proposed model and their adjustment will be investigated to improve the model's performance. The length of the feature vector in the character-based input is the first parameter whose value setting is checked. This parameter was tested with 30, 45, 60, and 75. The model's performance will be optimal considering the value of 45 for the feature vector length. The model with a vector length of 30 has obtained the lowest efficiency due to the reduction of the vector space coverage. The Tanh function was also chosen as the activation function of the character-based input CNN model, and the last layer of this model is the max pooling layer.

In another evaluation stage, several tests were implemented to obtain optimal values for the network parameters and the final model. The optimal values of network parameters are shown in Table 7.

**Table 7.** Hyperparameter setting in final model for NER task

| Parameter | Optimal Value |
|---|---|
| CNN Filter Size (3 Layer) | 15 |
| CNN Kernel Size (3 Layer) | 3 , 5 , 7 |
| CNN Window Size | 1 |
| CNN Activation Function | Tanh |
| CNN Pooling Function | Max Pooling |
| Bi-LSTM-1 unit | 275 |
| Bi-LSTM-2 unit | 100 |
| Bi-LSTM Dropout | 0.25 |
| Dropout | 0.50 |
| Learning Rate | 0.02 |
| Epoch | 200 |
| Optimization Function | Nadam |

Finally, after achieving the final structure for the architecture of the proposed model and repeating the experiments in the stage of fine-tuning the model's parameters and obtaining their optimal values, the model has been implemented with the final parameters and the number of epochs of 200.

i2b2 2010 dataset has three classes of problem, treatment, and test, and the final model is used for the NER task on this dataset. Table 8 shows the evaluation results of the final model for the named entity recognition of each of these classes separately. As can be seen, the model has performed best in the Treatment label and the lowest in the Test label.

Among the reasons for the weaker performance of the model in identifying the Test class compared to the other two classes, there are many abbreviations in laboratory factors, the similarity of some laboratory titles with prepositions in general texts and the appearance of entities of this class less than the other two classes in Clinical reports mentioned.

**Table 8.** Performance evaluation of final model for named entity recognition of i2b2 2010 dataset classes: problem, treatment, and test.

|  | Precision | Recall | F1-score | Support |
|---|---|---|---|---|
| Problem | 90.30 | 91.41 | 90.85 | 14831 |
| Treatment | 91.36 | 90.86 | 91.11 | 14058 |
| Test | 89.45 | 84.80 | 87.07 | 10069 |
| avg/total | 90.46 | 89.50 | 89.98 | 38958 |

Also, Table 9 shows the model's performance in NER on the i2b2 2010 dataset, separated by each class, compared to the work of [44]. According to the comparison, the proposed model performed better in identifying named entities in both Treatment and Problem classes.

**Table 9.** Comparative analysis of model performance in NER task for i2b2 2010 dataset against **[44]**.

|  | **BERT-Base on MIMIC** [44] | **BERT-Large on MIMIC** [44] | **Our Model** |
|---|---|---|---|
| Problem | 89.61 | 89.26 | **90.65** |
| Treatment | 88.09 | 89.14 | **91.21** |
| Test | 88.30 | **88.80** | 87.07 |

After numerous experiments, an ensemble model has also been implemented and evaluated. A combination of five single models with the same architecture and network parameters, each of which is initialized and trained with different random weights (random seeds). Finally, the test dataset was simultaneously processed and labelled in each separate model, and the results of all models were aggregated and evaluated using the voting process. More precisely, the label that received the most votes by the five models for each word in the dataset is selected as the final predicted label. Because each model starts to be trained with random weights, it is expected that combining these several models will improve the final result. As seen from the results in Table 10, this has increased the F1 measure by 0.07%.

Table 10. Comparison of ensemble model and individual models in NER task for i2b2 2010 dataset classes: Problem, Treatment, Test, and total f1-scores

|  | Model1 | Model2 | Model3 | Model4 | Model5 | Ensemble Model |
|---|---|---|---|---|---|---|
| Problem | 90.67 | 90.96 | 90.53 | 90.59 | 90.22 | 90.73 |
| Treatment | 91.42 | 91.26 | 91.11 | 91.27 | 91.35 | 91.28 |
| Test | 87.08 | 87.02 | 86.99 | 86.92 | 86.83 | 87.11 |
| Total | 90.01 | 89.89 | 89.88 | 89.98 | 89.78 | **90.11** |

For the final evaluation, Table 11 shows the performance results of this model in comparison with similar works in the field of the NER on the i2b2 2010 dataset.

Table 11. Analysis of F1-scores for the different NER models compared to our single and ensemble models.

| Model | F1-Score |
|---|---|
| Fully-connected LSTM–CRF(Ji et al. [40]) | 84.15 |
| RoBERTa-Large(Lewis et al. [42]) | 89.71 |
| Slide window model(Tang et al. [41]) | 89.25 |
| BiLSTM n-CRF(Nath et al. [24]) | 89.40 |
| RoBERTa pretrained on MIMIC(Yang et al. [45]) | 89.94 |
| BERT-Base pretrained on MIMIC(Si et al. [44]) | 89.55 |
| BERT-Large pretrained on MIMIC(Si et al. [44]) | 90.25 |
| Our Model(single) | 90.04 |
| Our Model(ensemble) | 90.11 |

One of the advantages of the proposed model is to have several inputs to process the text from different aspects and produce a vector of various features for each input word. According to the evaluations, the BlueBERT model, which is trained on the clinical report, had a stronger impact on improving the model's performance than any other word embedding model. An additional evaluation shows that the multi-channel CNN model when used in character-level embedding with a character pattern discovery approach makes the model more effective. Also, using word writing format embedding input focusing on identifying words with specific writing has affected the model's performance. Additionally, the novel pre-processing technique based on medical texts has improved the classification of multi-word medical terms through more accurate data labeling.

The final evaluations show that the proposed model has achieved a good performance in identifying named entities compared to similar works. The model proposed in [44] achieves better performance by using the BERT-Large architecture, which has more network parameters compared to BERT-Base. However, training this larger model requires more time and computational resources. The [47] model also demonstrates high accuracy on the i2b2 dataset, but uses a non-standard train/test split that differs from the canonical splits used in other work. Direct comparison is therefore difficult, since performance depends heavily on the train/test distribution.

## 5. Conclusion

In this paper, we proposed a model based on multi-layer neural network architecture for Named Entity Recognition. A rich feature vector for each word was generated using word-level embeddings, character embeddings, and word writing embeddings. Written, semantic, and relational features of that word are included in this feature vector, as well as the challenges involved in overcoming current clinical writing styles. According to the results of performance evaluation, using an advanced labeling method before introducing data to the model improves the accuracy and effectiveness of the model and identifies multi-word entities more accurately. Furthermore, the proposed model uses two layers of BiLSTM to identify multi-word entities and their relationships. The evaluations show that the proposed model has generally achieved better name entity recognition on the i2b2 dataset compared to similar works.

Further research in this area could be using a pre-trained model like the Elmo model in character input. The advantage of using this method for medical acronyms and some laboratory results that have limited characters. Also, words that have misspellings and one or more characters are different from the main word; in this case, the embedding produced in this model is very similar to embedding the original word without misspellings.

## 6. Appendix

Figure 8 illustrates how the F1 score for the BlueBERT model varies with different learning rate values. The learning rate was tested from 2e-5 to 5e-5 to find the optimal value that maximizes model performance. As shown, the F1 score steadily increases as the learning rate is raised from 2e-5 to 3e-5, reaching a peak F1 score of 89.3 with a learning rate of 3e-5. This indicates that a learning rate of 3e-5 results in the best performance for the BlueBERT model on this task.

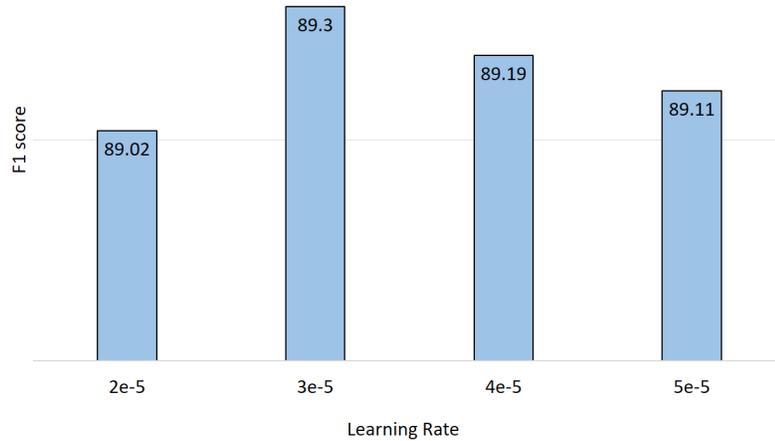

**Figure 8.** Analyzing the effect of various learning rate values on model F1-score.

Figure 9 demonstrates the impact of the number of training epochs on the F1 score for the model. The model was trained for between 3 to 10 epochs to determine the optimal number for maximizing performance. As illustrated, the F1 score steadily increases as the number of training epochs rises from 3 to 7, peaking at 89.3 after 7 epochs. This indicates that 7 training epochs are sufficient for the model to learn the patterns in the data without overfitting

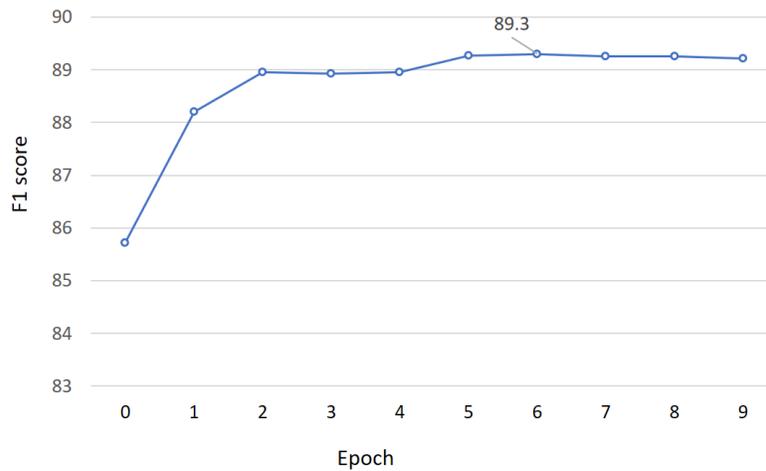

**Figure 9.** The impact of the number of training epochs on the F1-score of BlueBERT model.

Figure 10 shows the effect of varying the character-level feature vector length on model performance. The feature vector length was evaluated at 30, 45, 60, and 75 to determine the optimal value. As depicted, the F1 score increases steadily from 0.63 to 0.69 as the vector length is expanded from 30 to 45 characters. This demonstrates that a vector length of 45 characters provides the model with sufficient contextual information to effectively learn semantic representations. However, further lengthening the vector to 60 and 75 characters leads to a decrease in F1 score. The decline suggests that longer vectors introduce excessive noise and redundancy, deteriorating the model's ability to distill useful features. Overall, a character-level feature vector length of 45 results in the highest F1 score of 0.69, indicating it provides the best balance of coverage and conciseness for this model.

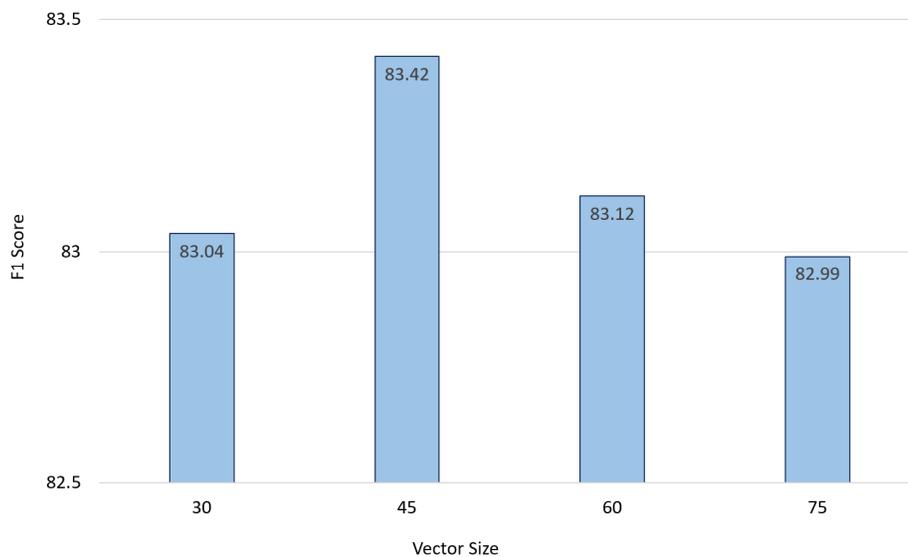

**Figure 10.** Impact of different character-level vector sizes on model's F1-score.

Figure 11 shows the F1 score progression over 200 training epochs for the finalized named entity recognition model on the i2b2 2010 dataset. In the initial epoch, the model starts at an F1 score of approximately 0.84, indicating a decent initial performance. During the first few epochs, there is a sharp upward surge in F1 score, rapidly increasing to around 0.87. This swift improvement suggests the model is quickly learning underlying patterns in the early phase of training. Following this initial spike, the F1 score continues trending upward but at a more gradual, stable rate as training progresses. This slower improvement means the model is still getting better at recognizing named entities, but learning is becoming harder.

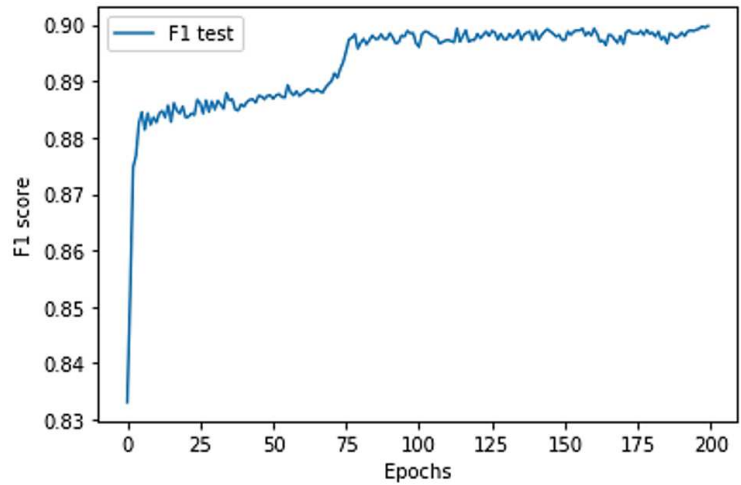

**Figure 11.** The impact of the number of training epochs on the F1-score of final model